\documentclass[10pt,twocolumn,letterpaper]{article}
\usepackage[pagenumbers]{cvpr}
\def\paperID{arXiv preprint}
\def\confName{CVPR}
\def\confYear{2027}
\title{ScheNav: Schematic-Guided Object Goal Navigation}
\author{%
\begin{tabular}{c}
Guichen Huang\textsuperscript{1,2} \quad Chenchen Jing\textsuperscript{3} \quad Che Sun\textsuperscript{2,*} \quad
Yuwei Wu\textsuperscript{1,2,*} \quad Yunde Jia\textsuperscript{2,1}\\
\textsuperscript{1}Beijing Institute of Technology \quad \textsuperscript{2}Shenzhen MSU-BIT University \quad \textsuperscript{3}Zhejiang University of Technology
\end{tabular}%
}

\definecolor{cvprblue}{rgb}{0.21,0.49,0.74}
\usepackage[pagebackref,breaklinks,colorlinks,allcolors=black]{hyperref}

\begin{document}

\maketitle
\begin{abstract}

Object goal navigation requires an embodied agent to locate and reach target objects in complex environments.
Humans often rely on low-cost yet informative schematic maps, such as rough sketches or directional annotations.
We develop a workflow for schematic-guided embodied navigation with two components: a training-data generation pipeline and ScheNav, a closed-loop navigation framework.
Our lightweight data-generation pipeline turns 2D floorplans into navigation training environments, yielding Furnished-RPLAN, a dataset of over 33k furnished layouts with paired navigation episodes and noise-controllable schematic maps.
ScheNav turns rough schematic guidance into exploration and stopping decisions by using SchemaField, a learned spatial module that fuses the guidance with accumulated observations and predicts spatial value fields.
The framework updates these fields as exploration proceeds, with trajectory-based training and a cross-frame consistency objective supporting coherent guidance across changing viewpoints.
ScheNav learns to use schematic guidance entirely from simulated floorplans and transfers this ability to unseen embodied environments.
Experiments show improved navigation success and path efficiency across schematic maps, goal ellipses, and pointed directions, together with tolerance to the tested schematic perturbations and omissions.
Together, these results demonstrate that rough schematic guidance offers a practical way to improve embodied navigation.

\end{abstract}

\section{Introduction}

\begin{figure}[t]
    \centering
    \includegraphics[width=0.95\columnwidth]{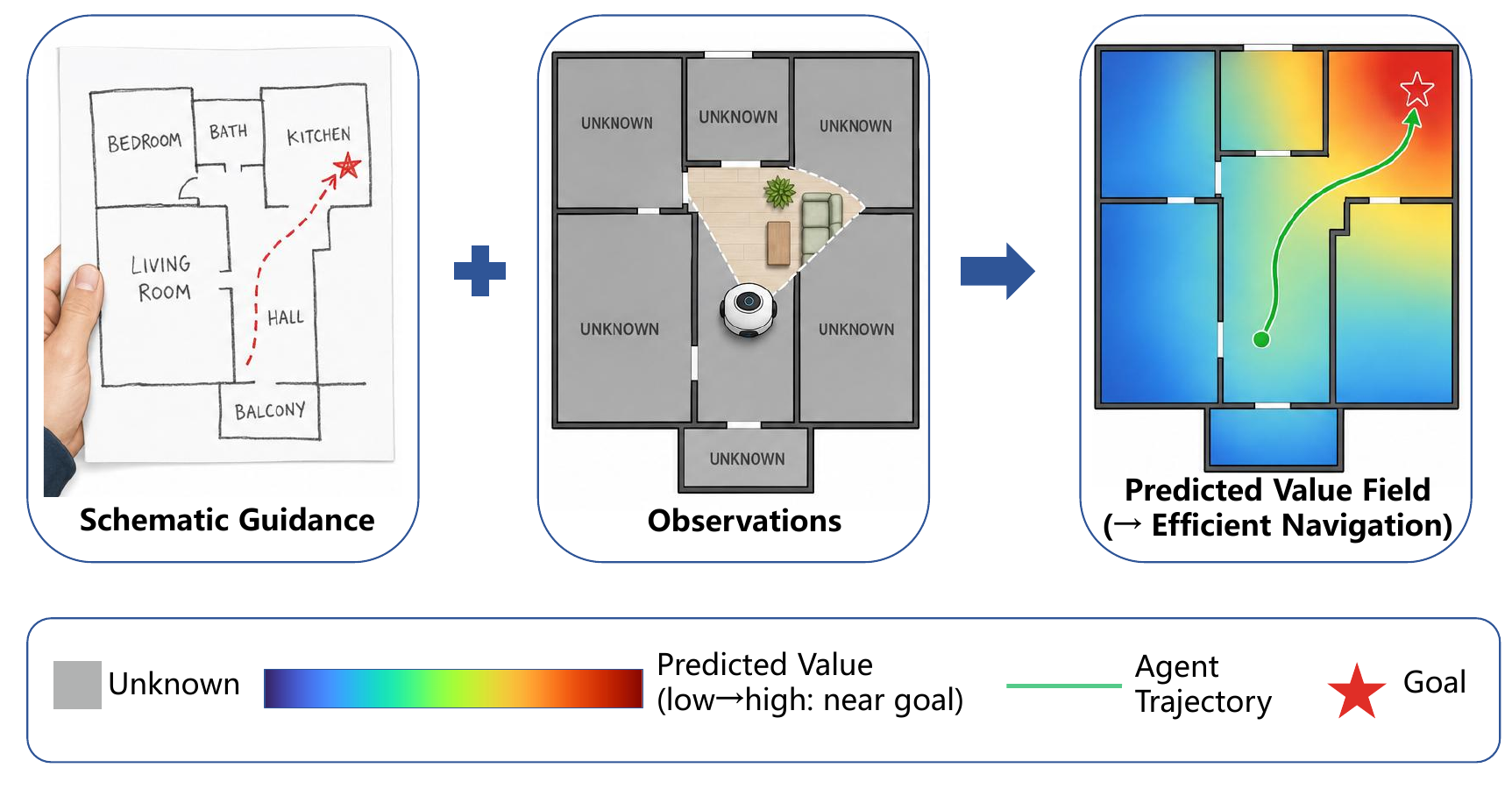}
    \caption{Overview of schematic-guided navigation. A human provides imperfect schematic guidance, such as a rough sketch or a pointed direction. ScheNav grounds the guidance in the agent's observations and predicts a value field that rises toward the goal. Guided by the predicted field, the agent reaches the goal along an efficient route in an environment it has barely explored.}
    \label{fig:teaser}
\end{figure}

Object goal navigation requires an embodied agent to locate and reach an instance of a specified object category in an unseen environment \cite{batra2020objectnav, zhu2017target}.
Recent map-free approaches, from modular semantic exploration to zero-shot agents built on large language and vision-language models \cite{chaplot2020object, gadre2023cows, zhou2023esc, yokoyama2024vlfm, yin2025unigoal}, accomplish this from egocentric observations alone and generalize encouragingly to novel scenes.
Meanwhile, map-based methods that plan on accurate global maps reach goals with high reliability, since a faithful map reduces navigation to well-understood localization and path planning \cite{thrun2005probabilistic, cadena2016past}.
Each line of work thus holds an advantage the other lacks.
Map-free agents pay no cost for map acquisition, while map-based agents enjoy strong global guidance about where to go.
We explore whether rough schematic guidance can help embodied agents navigate without requiring an accurate global map.

Humans routinely provide and use exactly this kind of guidance \cite{tversky1993cognitive}.
When searching for a room in an unfamiliar building, we rarely demand a precise map.
A rough sketch of the layout or a pointed direction is usually enough to focus the search.
Such schematic guidance can be produced with little effort and can convey useful layout and goal cues despite distortion and omission.
For example, marking a destination on a rough floorplan can help someone find a room without specifying an exact route.
Figure~\ref{fig:teaser} illustrates this setting, where the agent, given a rough sketch and local observations, is steered toward the goal.
Can the agent reliably convert such imperfect schematic guidance into effective closed-loop navigation decisions?

We develop a workflow for schematic-guided embodied navigation with two components: a training-data generation pipeline and ScheNav, a closed-loop navigation framework.
The workflow starts with 2D floorplans, from which we generate environments for learning how to use schematic guidance.
A vision-language model furnishes RPLAN floorplans \cite{wu2019data} and extracts machine-readable furniture layouts, which are processed against the original floorplan geometry.
From these layouts, we generate navigation episodes, simulated local observations, ground-truth spatial fields, and paired schematic maps with controllable noise and incompleteness.
Environment generation and navigation training take place entirely in 2D, without constructing 3D scenes or collecting embodied training trajectories.
This lightweight pipeline produces Furnished-RPLAN, a dataset of over 33k furnished layouts, and supports both training and controlled evaluation of schematic-guided navigation.

ScheNav turns the resulting training resources into a framework for using schematic guidance during navigation.
In each episode, the agent receives a goal category and guidance in the form of a schematic map, goal ellipses, or a pointed direction.
As the agent moves, its observations accumulate into a map of the area explored so far, which serves as its \emph{map memory}.
SchemaField, the only learned model in the framework, fuses the guidance with this observed map and predicts two spatial value fields.
The \emph{distance field} scores locations by their navigable proximity to the goal, while the \emph{corridor field} highlights efficient routes toward it.
The schematic is treated as a coarse prior to be interpreted alongside observations, rather than an accurate map on which to plan directly.
Fixed rules read the fields to guide exploration and verify goal candidates.
During exploration, the agent selects the \emph{frontier} with the highest corridor score \cite{yamauchi1997frontier}.
The \emph{Goal Verification} module checks whether a detected instance of the goal category maintains a high distance-field value over consecutive steps.
When this condition is met, the agent commits to a final approach and stops on arrival.
A single SchemaField supports the different guidance forms without retraining.
ScheNav attaches to an existing navigation system as a guidance layer, leaving perception, mapping, and low-level motion to its host.

Effective schematic guidance must remain coherent as the agent moves and new observations become available.
In our initial implementation, adding a field predictor trained on individual observation snapshots did not improve closed-loop navigation, despite producing plausible per-frame predictions.
Snapshot supervision, commonly used in learned spatial prediction for navigation \cite{ramakrishnan2022poni, georgakis2022learning}, does not by itself enforce agreement across viewpoints.
We observed that predictions for the same place could disagree after transformation into the world frame, causing successive frontier decisions to undo earlier progress; we call this failure \emph{cross-frame inconsistency}.
SchemaField is therefore trained on simulated navigation trajectories, with a cross-frame consistency objective that aligns predictions across viewpoints.
The corridor field further concentrates value along near-shortest paths, turning small differences in goal proximity into clearer frontier choices.
These designs allow the framework to update its guidance as exploration proceeds while maintaining coherent decisions across changing viewpoints.

We evaluate this workflow by training ScheNav entirely on simulated floorplans and deploying it in unseen embodied environments without embodied training data.
Integrated into UniGoal \cite{yin2025unigoal}, ScheNav uses schematic guidance to improve navigation success and path efficiency on the instance-level HM3D benchmark.
Experiments cover schematic maps, goal ellipses, and pointed directions, examining how different forms of guidance support navigation.
Controlled perturbations on Furnished-RPLAN further test tolerance to schematic noise and missing information.
Component ablations examine how trajectory-based training, cross-frame consistency, corridor prediction, and Goal Verification contribute to closed-loop performance.
A supplementary demonstration uses a genuinely hand-drawn map as input to the navigation pipeline.
Together, these experiments examine both the practical value of schematic guidance and the framework designs needed to use it effectively.

Our main contributions are summarized as follows:
\begin{itemize}
    \item We develop a workflow for schematic-guided embodied navigation that connects lightweight 2D training-data generation with a closed-loop navigation framework, providing a practical route for incorporating rough schematic guidance into embodied navigation.

    \item We construct Furnished-RPLAN, a dataset of over 33k furnished layouts with paired navigation episodes, ground-truth spatial fields, and noise-controllable schematic maps, through a pipeline that operates entirely in 2D without requiring 3D scene construction or embodied trajectory collection.

    \item We introduce ScheNav, a framework that converts schematic guidance and accumulated observations into exploration and goal-verification decisions. Trajectory-based training and cross-frame consistency support coherent guidance, and experiments evaluate transfer to embodied navigation across multiple guidance forms and controlled perturbations.
\end{itemize}

\section{Related Work}
\begin{figure*}[t]
    \centering
    \includegraphics[width=0.98\textwidth]{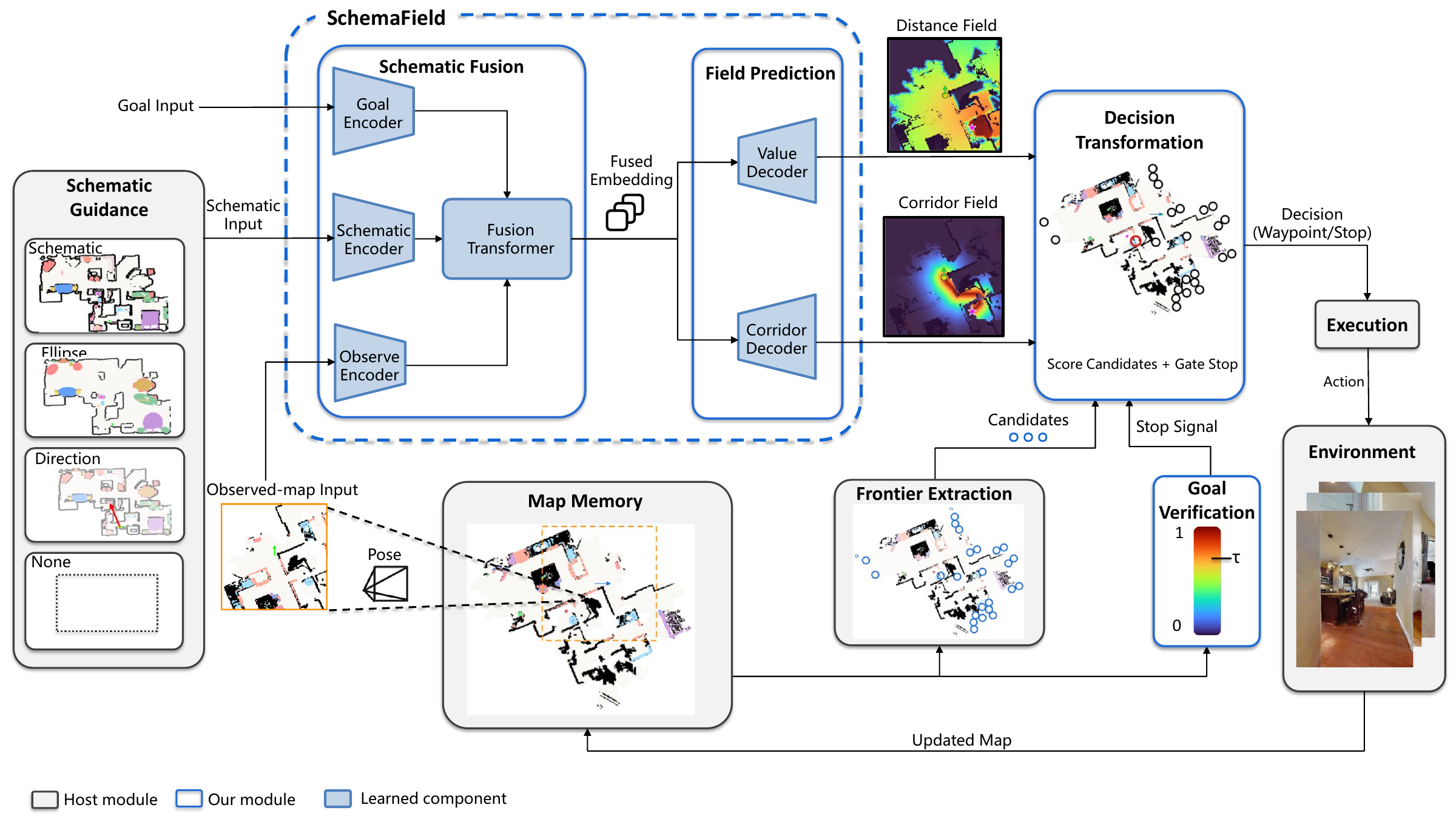}
    \caption{The ScheNav framework. Schematic guidance at any level, a schematic map, ellipses, a pointed direction, or none, is encoded by SchemaField, the only learned component our framework introduces: a schematic-fusion transformer fuses the guidance with the observed map from the map memory, and a distance decoder and a corridor decoder predict the two value fields from the fused embedding in a single forward pass. Two rule-based components read the fields. The Goal Verification module compares the distance-field value at goal instances detected in the map memory against a threshold $\tau_{\mathrm{stop}}$, and the decision transformation scores frontier candidates with the corridor field and gates stopping, emitting a waypoint or a stop. Map memory, frontier extraction, and execution are supplied by the host navigation system, and each new observation closes the loop.}
    \label{fig:architecture}
\end{figure*}

\noindent\textbf{Object Goal Navigation.}
Object goal navigation asks an embodied agent to find and reach an instance of a given object category in a previously unseen environment \cite{batra2020objectnav, zhu2017target}.
Modular approaches build a semantic map online and select exploration targets on it \cite{chaplot2020learning, chaplot2020object, ramakrishnan2022poni}, end-to-end policies learn navigation directly from raw observations at scale \cite{wijmans2020ddppo, mezghani2021memory}, and zero-shot methods score exploration directions with the commonsense knowledge of large language and vision-language models \cite{majumdar2022zson, gadre2023cows, zhou2023esc, yu2023l3mvn, yokoyama2024vlfm, yin2025unigoal}, with general-purpose frameworks unifying multiple goal modalities \cite{larsson2024goat, shah2023gnm, shah2023vint}.
These methods must infer where to search from egocentric observations or general-purpose priors.
Our workflow complements this setting by introducing low-cost, human-provided spatial guidance and grounding it in the observations accumulated by the agent.

\noindent\textbf{Navigation with Human Spatial Guidance.}
Using human-provided spatial guidance, such as sketches, route maps, and verbal cues, is a long-standing goal in robot navigation \cite{skubic2004sketch, tellex2011understanding}.
Early work localizes and navigates directly in hand-drawn sketches with unknown local metric distortions \cite{boniardi2015ecmr, boniardi2016icra}.
Language-based guidance has also proven effective, ranging from dialogue-based clarification \cite{thomason2020vision} to Vision-and-Language Navigation with sequential instructions \cite{anderson2018vision, fried2018speaker, ku2020room}.
More recently, HAM-Nav \cite{tan2025mobile} interprets hand-drawn maps with a vision-language model under scale distortions and missing landmarks, and FloNa \cite{li2025flona} shows that accurate architectural floorplans provide strong navigation priors.
This line of work, from sketch localization to floorplan priors, generally treats the provided map as the geometric space in which the agent localizes, plans, or reasons.
We instead use the schematic as a weak spatial prior that is fused with the agent's observations.
ScheNav encodes diverse schematic forms into a shared representation and grounds them in the partial map the agent has actually built, allowing the same framework to support schematic maps, ellipses, pointed directions, and no schematic input.
A hand-drawn sketch can enter this interface after conversion, while its effect is continuously modulated by the quality of the guidance and the evidence accumulated during navigation.

\begin{figure*}[t]
    \centering
    \includegraphics[width=0.9\textwidth]{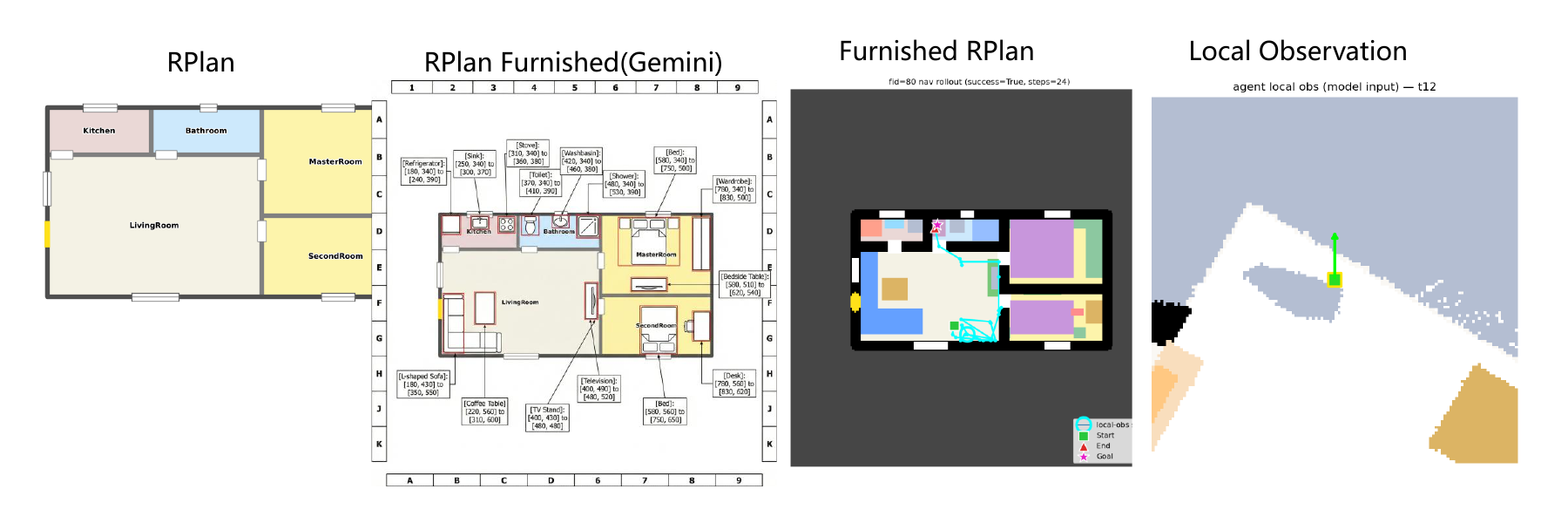}
    \caption{Furnished-RPLAN construction pipeline. Starting from an RPLAN vector floorplan, a vision-language pipeline proposes and extracts a machine-readable furniture layout. The resulting furnished environment is used to generate navigation episodes, paired schematic guidance, and the local observations consumed by ScheNav.}
    \label{fig:dataset_pipeline}
\end{figure*}

\noindent\textbf{Spatial Priors and Value Prediction for Navigation.}
Semantic priors learned from large-scale 3D environments, such as room layouts and object co-occurrences, improve navigation in unseen scenes \cite{yang2018visual, mousavian2019visual, mayo2021visual}, supported by simulators and datasets including Gibson \cite{xia2018gibson}, Matterport3D \cite{chang2017matterport3d}, HM3D \cite{ramakrishnan2021hm3d}, and Habitat \cite{savva2019habitat, szot2021habitat}.
Closest to our formulation, active exploration methods predict potential functions or semantic maps from the partial map and score frontiers accordingly \cite{chaplot2020learning, ramakrishnan2022poni, savinov2018semi, georgakis2022learning}.
However, when these predictors are supervised on individual snapshots, their outputs are not constrained to remain consistent across viewpoints, which can undermine closed-loop decisions.
Our trajectory-based training and cross-frame consistency objective align predicted fields in the world frame under exact relative pose transforms, directly constraining the quantity used for navigation decisions.
Complementary to existing 3D embodied priors, RPLAN provides large-scale structured floorplan data \cite{wu2019data, nauata2020house, armeni20193d}.
Our training-data generation pipeline furnishes these floorplans with machine-readable layouts, paired navigation episodes, and schematic guidance, allowing us to train spatial prediction entirely on simulated environments and transfer the resulting framework to embodied navigation. Figure~\ref{fig:dataset_pipeline} illustrates this pipeline, from an RPLAN floorplan and VLM-generated furnishing to the machine-readable environment and local observation consumed by the model.

\section{Method}

\subsection{Framework Overview}
At each time step $t$, the agent receives a semantic observation $o_t$, its pose $p_t$, and a goal category $g$, together with optional schematic guidance $s$: a schematic map optionally annotated with start and goal markers, goal ellipses alone, or merely a pointed direction. The guidance is imperfect. It may be noisy, incomplete, spatially distorted, partially annotated, or entirely absent, and it is therefore treated as a weak human spatial prior rather than as metrically aligned supervision.

ScheNav organizes navigation as a closed loop (Figure~\ref{fig:architecture}). The \emph{map memory} accumulates the observation history into the partial map $m_t$ and, through a standard frontier-extraction routine, supplies the current frontier candidates and the visible instances of the goal category. \emph{SchemaField}, the only learned component our framework introduces, predicts two value fields over the explored area,
\begin{equation}
(V_t,\, C_t) = f_{\theta}(s,\, m_t,\, p_t,\, g),
\end{equation}
where the \emph{distance field} $V_t$ scores every location by its navigable proximity to the goal and the \emph{corridor field} $C_t$ concentrates value along efficient routes toward it. The \emph{decision transformation} and the \emph{Goal Verification} are two rules with no learned parameters: the former selects the next frontier by its corridor score, and the latter commits to stopping once the distance-field value at a detected goal instance stays persistently high. The \emph{execution} component turns each decision into motion and feeds new observations back into the map memory, closing the loop.

The loop dictates the central requirement on the learned model. Because SchemaField is re-queried at every step from a different pose, its outputs must agree in the world frame across queries. This consistency requirement is part of the model's contract rather than an optional refinement, and it shapes both the training setup and the decision target described below. The challenge throughout is not to reconstruct an accurate global map, but to convert weak schematic guidance into predictions that remain reliable over an entire episode while degrading gracefully as the guidance weakens.

\subsection{Host Navigation Interface}

Both components in this section are supplied by the host navigation system and are specified by interface rather than by a particular implementation; a lightweight floorplan simulator instantiates them during training, and an existing navigation system instantiates them at deployment.

\paragraph{Map memory.}
The map memory accumulates the observation history into the partial map $m_t$ and answers two queries used by the rest of the framework: the current frontier candidates, obtained by a standard frontier-extraction routine, and the visible instances of the goal category, obtained by a non-learned connected-component detection on the egocentric semantic observation, which the Goal Verification evaluates to issue the stop signal.

\paragraph{Execution.}
The execution component takes the selected frontier or homing target, invokes the underlying navigation stack to move toward it, and updates the map memory with the new observations.

\subsection{SchemaField}

At each step SchemaField receives an egocentric semantic crop of the memory, the schematic guidance, the agent pose, and the goal category. Whenever schematic input is present, it is coarsely registered into the egocentric observation frame by the annotated start position and the agent heading, and the remaining distortion is intentionally left for the model to absorb. Structural schematic content and optional annotations are encoded separately, so that the same model can consume a schematic map with or without a goal marker, goal ellipses, a noisy pointed direction, or no schematic input at all.

All inputs are mapped into a unified token space and processed by a single joint Transformer,
\begin{equation}
z_f = \mathrm{SelfAttn}([\,z_g;\, z_s;\, z_o\,]),
\end{equation}
where $z_o$, $z_s$, and $z_g$ denote observation, schematic, and goal tokens. Every token attends to every other token, allowing observation tokens to query schematic structure directly, schematic tokens to ground themselves in observed geometry, and the goal token to condition both.

From the fused representation SchemaField decodes the two fields over the explored area. The distance field $V_t$ is supervised as a geodesic proximity map, in which values increase as locations become closer, in navigable distance, to the target region implied by the goal and the schematic prior. Proximity alone leaves small margins between competing frontiers, so exploration decisions are read from a second target. The corridor field $C_t$ has a supervision target $\bar{C}_t$ uniquely defined by the environment geometry and the agent's current position,
\begin{equation}
\bar{C}_t(x)=\exp\!\Big(\!-\frac{d_a(x) + d_g(x) - d^{*}}{\tau_c}\Big)\cdot \sigma\!\Big(\frac{d^{*} - d_g(x)}{\kappa}\Big),
\end{equation}
where $d_a(x)$ and $d_g(x)$ are geodesic distances from location $x$ to the agent and to the goal, $d^{*}$ is the shortest agent-to-goal geodesic distance, $\sigma$ is a sigmoid gate, and $\tau_c$ and $\kappa$ are scale hyperparameters (values in the supplementary material). The first factor concentrates mass on locations lying on a near-shortest path between agent and goal, giving a large decision margin. The forward gate suppresses the geometrically symmetric branch behind the agent, which we found to be the dominant source of residual direction flips.

The supervision of both fields is aligned with the input condition. When the goal marker is given, the target is the geodesic field of the annotated instance. When only the goal category is known, the target is the category-level field of the nearest instance of that class, and the corridor target is redefined. During training, \emph{condition dropout} exposes the same model to all guidance conditions so that deployment can switch among them without retraining.
We deliberately keep the two fields, rather than a low-level action distribution, as the interface between prediction and decision. This makes SchemaField easy to attach to an existing navigation framework, replacing only the spatial decision steps while leaving perception, mapping, and local planning unchanged.

\subsection{Decision Transformation}

The decision transformation contains no learned parameters. At each decision point it scores the frontier candidates $\mathcal{U}_t$ queried from the map memory with the corridor field and steers the agent toward the highest-scoring one, $u^{*} = \arg\max_{u \in \mathcal{U}_t} C_t(u)$, while stopping is governed by a dedicated gate.

\paragraph{Goal Verification.}
The gate queries the map memory for the visible instances of the goal category and samples the distance field at their locations. When the sampled value exceeds a threshold $\tau_{\mathrm{stop}}$ for two consecutive steps, the agent commits, switches from exploration to homing on the best-valued instance, and stops on arrival. The commitment is never reverted, while the homing target keeps updating to the currently best instance. Because the absolute scale of the predicted field differs across input conditions, $\tau_{\mathrm{stop}}$ is calibrated per condition on held-out simulated floorplans and then frozen before embodied evaluation.

\subsection{Trajectory-Consistent Training}

\paragraph{Failure mode.}
A distance field trained on independent observation snapshots can be accurate in every frame yet still fail to drive navigation. Because the field is predicted in the egocentric frame, two predictions made before and after a large rotation must agree once re-expressed in the world frame, but with single-snapshot training nothing enforces this. Empirically, after the agent turns by $90^{\circ}$ or more, the world-frame correlation between consecutive predictions drops to $-0.15$, and the movement direction implied by consecutive frontier decisions flips in $45.6\%$ of cases. In a closed navigation loop these flips cancel each other's progress, which explains why a field that looks good per frame does not convert into navigation success.

\paragraph{Sequence training.}
Instead of sampling independent snapshots, we generate training data along simulated navigation trajectories. The observed area grows monotonically as the agent explores, and the heading varies naturally along the path. Each trajectory contributes a sequence of (observation, schematic, target-field) tuples whose statistics match what the model encounters at deployment, including partially explored maps and arbitrary orientations.

\paragraph{Cross-frame consistency objective.}
On top of sequence sampling, we add an explicit consistency loss. For pairs of frames of the same underlying state under different headings, and for adjacent snapshots along a trajectory, we align the two predicted fields with the exact relative rigid transform between the two poses and penalize disagreement on the jointly visible region. This directly optimizes the property that closed-loop navigation needs: predictions expressed in a common frame must agree regardless of the agent's viewpoint. Together, sequence training, the consistency objective, and the corridor target reduce the direction-flip rate from $45.6\%$ to $10.1\%$ and lift the post-turn world-frame correlation from $-0.15$ to $+0.80$. We ablate each design choice in the experiments.

\subsection{Furnished-RPLAN Benchmark}

We build \emph{Furnished-RPLAN}, a large-scale benchmark of furnished indoor environments derived from the RPLAN floorplan dataset \cite{wu2019data}. Each vector floorplan provides exact occupancy, semantics, and geodesic distance fields; on top of it, a two-stage vision-language pipeline produces a machine-readable furniture layout normalized to roughly forty Habitat furniture categories. We then generate navigation episodes and paired schematic maps with controllable noise and incompleteness. 
The final model is trained on $33{,}992$ furnished layouts, with layouts held out by floorplan ID for validation and threshold calibration.
Full construction details and pipeline visualizations are deferred to the supplementary material.

\section{Experiments}

\subsection{Experimental Setup}


\paragraph{Environments and protocols.}
We evaluate ScheNav on embodied navigation in Habitat \cite{savva2019habitat, szot2021habitat} with HM3D scenes \cite{ramakrishnan2021hm3d}, under the text-goal benchmark introduced by UniGoal \cite{yin2025unigoal}: 1000 episodes with the official success criterion, each targeting a specific object instance that baseline methods identify through a free-form textual description; we adopt the protocol unchanged. ScheNav is integrated into UniGoal, replacing its frontier selection and stopping decisions; the replaced decisions are driven only by the goal category and the schematic guidance, while the rest of the UniGoal pipeline, including its text interface, is left untouched, so all arms and baselines run the same episodes under the same criterion. We run the official UniGoal implementation; its LLM serving stack is detailed in the supplementary material. 
This benchmark is distinct from, and substantially harder than, category-level ObjectNav, on which the same baselines report much higher numbers.
Controlled component analyses use a frozen 122-episode, 36-scene subset of the official episodes. 

\paragraph{Metrics.}
For navigation we report Success Rate (SR), Success weighted by Path Length (SPL), soft-SPL, and distance to success (DTS). For frontier quality we report how often the selected frontier matches the ground-truth-optimal choice and the ground-truth geodesic distance of the selected frontiers. For cross-frame consistency we report the direction-flip rate, how often the movement direction implied by consecutive decisions reverses, and the world-frame correlation between consecutive predictions after large rotations. Field-quality metrics used for model selection and training hyperparameters are provided in the supplementary material.

\subsection{Main Results}

\paragraph{Comparison with the state of the art.}
Table~\ref{tab:sota} places ScheNav in the context of published methods under the instance-level HM3D benchmark of UniGoal described above, including the recent Context-Nav \cite{jang2026contextnav}, which holds the best published success rate. The comparison is not information-equal: baselines resolve a textual goal description while our agent receives schematic guidance, so we report four guidance levels, from the schematic map down to a single noisy pointed direction and to no guidance at all, quantifying what each level buys over the strongest text-driven systems. With the schematic map, ScheNav reaches $61.4\%$ success and $36.3\%$ SPL, $35.2$ and $24.9$ points above the best published numbers, while trained purely on simulated floorplan data. 
The gradient itself is informative: goal ellipses alone already recover most of the gain 
($60.4\%$), suggesting that a rough indication of where the goal lies matters more than detailed layout structure, and even the direction-only condition surpasses every text-driven baseline in success rate. The SPL gap tells the same story from the efficiency side: guided runs do not merely succeed more often, they reach the goal along far shorter paths.

\begin{table}[t]
\centering
\begin{tabular}{lcc}
  \toprule
Method & SR $\uparrow$ & SPL $\uparrow$ \\
\midrule
VLFM \cite{yokoyama2024vlfm} & 14.9 & 9.3 \\
GOAT \cite{larsson2024goat} & 17.0 & 8.8 \\
PSL \cite{chen2024prioritized} & 26.0 & 10.2 \\
UniGoal \cite{yin2025unigoal} & 20.2 & 11.4 \\
Context-Nav \cite{jang2026contextnav} & 26.2 & 9.1 \\
\midrule
Ours (no schematic input) & 21.9 & 5.8 \\
Ours (direction only) & 34.6 & 19.3 \\
Ours (ellipse) & 60.4 & 35.5 \\
\textbf{Ours (schematic map)} & \textbf{61.4} & \textbf{36.3} \\
  \bottomrule
\end{tabular}
\caption{Comparison with the state of the art on UniGoal's instance-level HM3D benchmark (1000 episodes; SR and SPL in percent). Baseline numbers follow Context-Nav \cite{jang2026contextnav}. Baselines receive a textual goal description; all ``Ours'' rows use the unified 34k model with guidance at the indicated level.}
\label{tab:sota}
\end{table}

\paragraph{Component progression.}
Table~\ref{tab:habitat_nav} traces how each component contributes on the frozen 122-episode set, where ScheNav replaces UniGoal's frontier selection and stopping decisions, making the comparison against the unmodified host apples-to-apples. All progression rows are trained on the same 800 layouts, so their comparisons isolate the components at a fixed data scale. The first two rows are deliberately incomplete variants: adding a single-frame distance field does not improve success ($15.6\%$ vs.\ $17.2\%$), and adding a retrained Goal Verification predictor on top of that field further reduces performance to $12.3\%$. These results show that appending individual components without coherent trajectory training is insufficient. Sequence training, the consistency objective, and the corridor head lift success to $23.8\%$; adding Goal Verification to this coherent navigation framework then raises it to $45.9\%$. The final unified model adds condition dropout and condition-aligned supervision and scales training to 16k layouts, reaching $54.9\%$ success against the $17.2\%$ baseline; enlarging the training set to the full 34k layouts adds a further $9.9$ points ($64.8\%$), showing that the transfer keeps improving with floorplan diversity. All our variants are trained purely on simulated floorplan data with no embodied training data.

\begin{table}[t]
\centering
\small
\setlength{\tabcolsep}{3.5pt}
\resizebox{\columnwidth}{!}{%
\begin{tabular}{lcccc}
  \toprule
Method & SR $\uparrow$ & SPL $\uparrow$ & sSPL $\uparrow$ & DTS $\downarrow$ \\
\midrule
UniGoal (baseline) & 17.2 & 9.5 & 21.3 & 5.65 \\
\midrule
\shortstack[l]{+ single-frame field\\(incomplete)} & 15.6 & 9.4 & 22.5 & 5.26 \\
\shortstack[l]{+ Goal Verification on single-frame\\field (incomplete)} & 12.3 & 8.5 & 23.3 & 6.16 \\
+ seq.\ + consist.\ + corridor & 23.8 & 15.5 & 28.7 & 4.30 \\
+ Goal Verification (complete) & 45.9 & 27.1 & 38.7 & 2.53 \\
Ours (all, $\sim$16k) & 54.9 & 35.0 & \textbf{48.6} & 1.75 \\
Ours (all, $\sim$34k) & \textbf{64.8} & \textbf{35.5} & 45.1 & \textbf{1.42} \\
  \bottomrule
\end{tabular}%
}
\caption{Component progression on the frozen Habitat set (122 episodes, 36 scenes; SR, SPL, and sSPL (soft-SPL) in percent, DTS in meters). Rows marked ``+'' progressively modify UniGoal \cite{yin2025unigoal} and are trained on 800 layouts. The first two rows are incomplete combinations included to show that adding components independently can hurt; the complete Goal Verification row follows trajectory-consistent field training. The last two rows are the unified model trained on $\sim$16k and $\sim$34k layouts.}
\label{tab:habitat_nav}
\end{table}



\begin{table}[t]
\centering
\setlength{\tabcolsep}{4pt}
\begin{tabular}{lcc}
\toprule
Metric & Before & After (full) \\
\midrule
Direction-flip rate ($>90^{\circ}$) $\downarrow$ & 0.456 & \textbf{0.101} \\
World-frame corr.\ (post-turn) $\uparrow$ & $-0.15$ & \textbf{+0.80} \\
Frontier ranking corr.\ $\uparrow$ & 0.35--0.48 & \textbf{0.750} \\
Top-decision margin $\uparrow$ & $1\times$ & \textbf{3.4}$\times$ \\
\midrule
\multicolumn{3}{l}{\emph{Corridor ablation (flip rate $\downarrow$):}} \\
plain corridor ($\tau_c{=}60$) & \multicolumn{2}{c}{0.367} \\
sharpened ($\tau_c{=}30$) & \multicolumn{2}{c}{0.380} \\
+ forward gate ($\kappa{=}20$) & \multicolumn{2}{c}{\textbf{0.101}} \\
\bottomrule
\end{tabular}
\caption{Cross-frame consistency before and after the fix (paired counterfactual replay, 97 decisions) and the corridor ablation (direction-flip rate), measured on the component-progression generation.}
\label{tab:consistency}
\end{table}

\begin{figure*}[t]
\centering
    \includegraphics[width=0.98\textwidth]{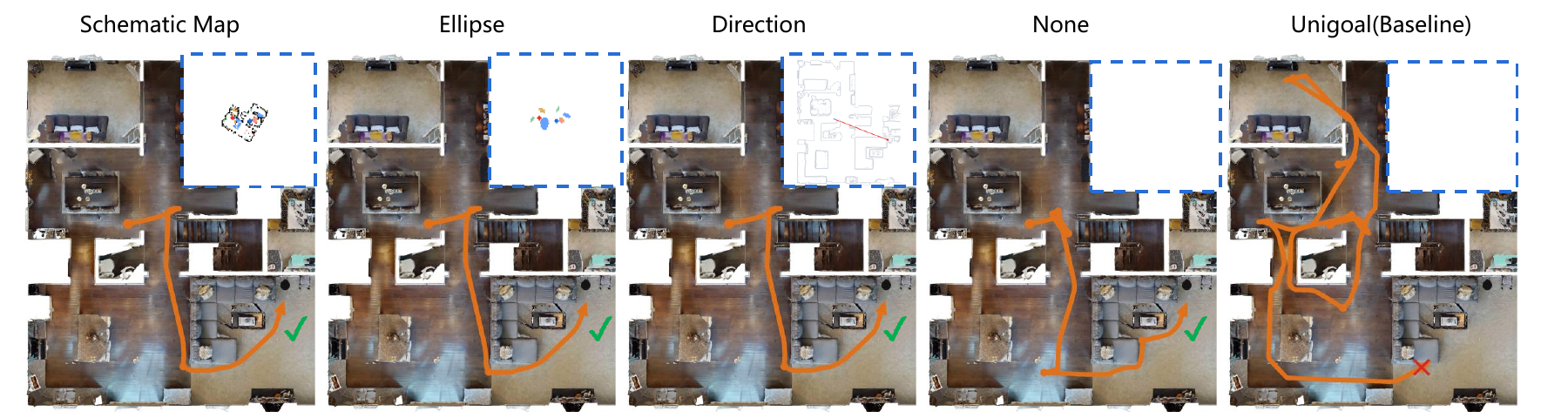}
    \caption{Closed-loop trajectories on one episode under decreasing guidance. The five panels show our four input conditions and unmodified UniGoal on the same episode, rendered on a textured top-down view of the scene. The dashed inset in each panel shows the schematic input the agent receives, and is blank for the no-guidance condition and the baseline; in the direction panel the floorplan behind the arrow is drawn only for illustration, and the model receives the direction vector alone. Orange curves trace the agent from start to stop, a green check marks a successful stop, and the red cross marks a failed stop. The path lengthens gracefully as guidance weakens, from the schematic map through the ellipse and direction conditions to no guidance at all, while the host stops incorrectly after prolonged, tangled exploration.}
    \label{fig:qualitative_rollout}
\end{figure*}

\subsection{Consistency and Decision Quality}

Does a field that looks better per frame make better closed-loop decisions? Table~\ref{tab:habitat_nav} already says no: single-frame training improves neither navigation success nor stopping, despite producing plausible snapshots. To isolate why, we replay competing models over the same 97 decision points on identical trajectories, so any difference in their decisions is attributable to the model alone (Table~\ref{tab:consistency}). The full model reduces the direction-flip rate from $45.6\%$ to $10.1\%$, turns the post-turn world-frame correlation from $-0.15$ to $+0.80$, raises the frontier-ranking correlation to $0.750$, and enlarges the top-decision margin by $3.4\times$.
Better frontier scoring alone is not the answer. When ScheNav replaces only UniGoal's frontier scoring, it picks the ground-truth-optimal frontier in $63.7\%$ of decisions against $25.5\%$ for the unmodified host, yet the single-frame variant still fails in closed loop: the per-decision gains are undone by cross-frame inconsistency. The corridor ablation in the lower block of Table~\ref{tab:consistency} makes the same point at the design level: merely sharpening the corridor does not reduce flips ($0.367$ vs.\ $0.380$), whereas the forward gate, which suppresses the geometrically symmetric branch behind the agent, cuts the flip rate to $0.101$; this design is shared by all subsequent models.

\subsection{Input Conditions and Fairness}
A natural concern is that comparing a schematic-guided agent against a map-free baseline is unfair, since the schematic input carries privileged information. Table~\ref{tab:sota} is thus best read as a dose--response curve rather than a leaderboard, showing how performance falls as the privilege is removed. Even a single noisy pointed direction surpasses the map-free baseline in success rate, and with no schematic input at all the framework still edges out its host ($21.9\%$ vs.\ $20.2\%$) at a lower SPL that reflects longer exploration. The zero-guidance arm is the purest probe of the learned model: what SchemaField has learned from floorplan training still extracts useful structure from the agent's own observations, paying off mainly at stopping, and that alone suffices to edge past the host. One design note: since no agent, human or learned, can make meaningful global decisions over a nearly empty map, decision authority under this condition is gated by map coverage, with the host leading early exploration and the fields taking over as observations accumulate. Since attaching the framework does not degrade its host even with zero privilege, the gains at higher guidance levels are attributable to the guidance itself. A subtler question arises when the input cannot identify a goal instance, so crediting only the annotated target is no longer fair; we leave a systematic treatment of category-level credit to future work.

\subsection{Qualitative Analysis}

Figure~\ref{fig:qualitative_rollout} compares closed-loop trajectories on a representative episode, with our four guidance levels and the unmodified host running the same episode. The input-condition gradient is directly visible: under the schematic-map, ellipse, or direction conditions, the agent proceeds to the goal along a nearly direct route, while the no-guidance arm takes a longer but still purposeful path. 
The host, in contrast, wanders through repeated revisits and ends with an incorrect stop. A demonstration in which a user hand-draws the schematic map and the agent navigates from it end to end is provided in the supplementary material.

\subsection{Failure Analysis and Limitations}

We classify all failed episodes of each guidance arm on the 1000-episode benchmark into three exclusive categories: \emph{wrong commits}, where the agent commits to an incorrect instance; \emph{late stops}, where the goal category is detected but the episode ends without a correct stop; and \emph{exploration misses}, where the goal is never observed. With a schematic map, failures concentrate in the stopping chain: $67.6\%$ are late stops, and a quarter of failures end within one meter of the success radius, while wrong commits account for $16.3\%$. As guidance weakens, wrong commits take over, reaching $62.1\%$ under direction-only guidance and $74.1\%$ with no guidance, driven by host-perception false positives compounded by same-class ambiguity. Excluding episodes ended by wrong commits raises the schematic-map success rate from $61.4\%$ to $65.5\%$, bounding the cost of host perception errors at four points.
Two directions follow. First, under strong guidance the residual bottleneck is stopping precision rather than spatial reasoning; a finer-grained terminal policy on top of the Goal Verification is a natural extension. Second, under weak guidance same-class instance ambiguity dominates, calling for instance-level reasoning beyond category cues.

\section{Conclusion}

We presented an end-to-end workflow for schematic-guided embodied navigation, combining a training-data generation pipeline with ScheNav, a closed-loop navigation framework.
ScheNav grounds imperfect schematic guidance in the agent's map memory and converts it into exploration and stopping decisions through a learned spatial module and fixed decision rules.
We also identified cross-frame inconsistency as a key failure mode when per-frame spatial prediction drives closed-loop navigation, and addressed it with trajectory-based training, a cross-frame consistency objective, and a geodesic-corridor decision target.
Trained purely on Furnished-RPLAN and integrated into UniGoal, the resulting framework raises the success rate of the underlying map-free host from 20.2\% to 61.4\% when given a schematic map, while even a single noisy pointed direction retains a substantial portion of the gain.
These results demonstrate that rough, human-provided spatial guidance can be made usable for closed-loop embodied navigation through a practical and reusable workflow.

\newpage

{\small
\bibliographystyle{ieeenat_fullname}
\bibliography{reference}
}

\end{document}